\documentclass[sigconf]{acmart}

\usepackage{natbib} 
\usepackage{booktabs}
\usepackage{comment}
\usepackage{graphicx}
\usepackage{subfig}
\usepackage{fancyhdr}
\usepackage{amsmath}
\usepackage{xurl}
\usepackage{adjustbox}
\usepackage{arydshln}
\usepackage{relsize}
\usepackage{wasysym}
\usepackage{multirow}
\usepackage[makeroom]{cancel}
\usepackage{scalerel,graphicx,xparse}
\usepackage{hyperref}
\usepackage{xcolor}
\usepackage{algorithm}
\usepackage{algpseudocode}
\usepackage{lscape}

\newcommand{\secref}[2][]{Section#1~\ref{sec:#2}}

\newcommand{\tabref}[2][]{Table#1~\ref{tab:#2}}

\AtBeginDocument{%
  \providecommand\BibTeX{{%
    \normalfont B\kern-0.5em{\scshape i\kern-0.25em b}\kern-0.8em\TeX}}}

\copyrightyear{2023}
\acmYear{2023}
\setcopyright{acmlicensed}\acmConference[WWW '23]{Proceedings of the ACM Web Conference 2023}{May 1--5, 2023}{Austin, TX, USA}
\acmBooktitle{Proceedings of the ACM Web Conference 2023 (WWW '23), May 1--5, 2023, Austin, TX, USA}
\acmPrice{15.00}
\acmDOI{10.1145/3543507.3583417}
\acmISBN{978-1-4503-9416-1/23/04}

\title{MetaTroll: Few-shot Detection of State-Sponsored Trolls with Transformer Adapters}

\author{Lin Tian}
\affiliation{%
  \institution{RMIT University}
  \city{Melbourne}
  \country{Australia}}
\email{lin.tian2@student.rmit.edu.au}

\author{Xiuzhen Zhang}
\affiliation{%
  \institution{RMIT University}
  \city{Melbourne}
  \country{Australia}}
\email{xiuzhen.zhang@rmit.edu.au}

\author{Jey Han Lau}
\affiliation{%
  \institution{The University of Melbourne}
  \city{Melbourne}
  \country{Australia}}
\email{jeyhan.lau@gmail.com}

\begin{document}


\begin{abstract}
State-sponsored trolls are the main actors of influence campaigns on social media and automatic troll detection is important to combat misinformation at scale. 
Existing troll detection models are developed based on training data for known campaigns (e.g.\ the influence campaign by Russia's Internet Research Agency on the 2016 US Election), and they fall short when dealing with {\em novel} campaigns with new targets. 
We propose MetaTroll, a text-based troll detection model based on the meta-learning framework that enables high portability and parameter-efficient adaptation to new campaigns using only a handful of labelled samples for few-shot transfer. We introduce \textit{campaign-specific} transformer adapters to MetaTroll to ``memorise'' campaign-specific knowledge so as to tackle catastrophic forgetting, where a model ``forgets'' how to detect trolls from older campaigns due to continual adaptation. 
Our experiments demonstrate that MetaTroll substantially outperforms baselines and state-of-the-art few-shot text classification models. Lastly, we explore simple approaches to extend MetaTroll to 
multilingual and multimodal detection. Source code for MetaTroll is available at: https://github.com/ltian678/metatroll-code.git
\end{abstract}



\maketitle

\section{Introduction}
State-sponsored trolls, along with bots, are the main actors in influence operations and misinformation campaigns on social media.
The 2016 US Presidential election for example, highlighted how foreign states can carry out mass influence campaign, and in this case by Russia's Internet Research Agency \cite{zhang2021assembling}.\footnote{\url{https://www.adelaide.edu.au/newsroom/news/list/2021/12/09/understanding-mass-influence-activities-is-critical}.}
In another context, \citet{broniatowski2018weaponized} studied how influence operations undermined the vaccine debate in public health. 
To help combat misinformation on social media, Twitter began releasing user accounts associated with state-sponsored influence activities. Our work uses this dataset for troll detection, and as such a {troll} in our context is a state-sponsored agent with links to an influence campaign.


Existing troll detection models focus on extracting signals from user online posts and user activities \citep{dlala2014trolls,cheng2015antisocial,atanasov2019predicting,de2013filtering,shafiei2022detection, addawood2019linguistic,alizadeh2020content}.
Recent neural approaches explore fusing different feature representations for troll detection, e.g.\ social structure, source posts and propagated posts such as retweets on Twitter~\cite{atanasov2019predicting,zannettou2019disinformation}.
These studies generally formulate the task as a standard supervised learning problem which requires a substantial amount of labelled data for known campaigns. 
As such, they are ill-equipped to detect { novel} campaigns sponsored by another state for a different target.

Meta-learning is a well-established framework for few-shot learning~\citep{finn2017model,vinyals2016matching,snell2017prototypical,garnelo2018conditional,requeima2019fast}.
The idea of meta-learning is to leverage the shared knowledge from previous tasks to facilitate the learning of new tasks.
In our context, meta-learning has the potential to quickly adapt a troll detection model to a new campaign via few-shot transfer, once it's meta-trained on a set of known campaigns. However, in a continual learning setting where the troll detection model needs to be constantly updated to work with new campaigns over time, the standard meta-learning framework suffers from catastrophic forgetting --- the problem where the model ``forgets'' how to detect trolls from the older campaigns as they are updated \citep{yap2021addressing,Xu+:2020}.

The main contributions of this work are as follows:
\begin{itemize}
\setlength\itemsep{0em}
\item We introduce a troll detection problem under a realistic setting where novel campaigns continually emerge, mimicking real-world events on social media platforms. 
\item We propose MetaTroll, a text-based meta-learning framework for troll detection, which includes a three-stage meta-training process that allows it to learn knowledge across different  campaigns for fast adaptation to new campaigns. 
\item MetaTroll tackles catastrophic forgetting by introducing \textit{campaign-specific} transformer adapters \cite{houlsby2019parameter} (i.e.\ each campaign has its own set of adapter parameters), 
which can ``memorise'' campaign-specific knowledge.
\item MetaTroll has the ability to to work with multiple languages (multilingual) by using a pretrained multilingual model and incorporate images  as an additional input (multimodal) by either encoding them using pretrained image classifiers or converting them into text via optical character recognition.
\item Large-scale experiments on a real-world dataset of 14 Twitter campaigns showed the superior performance of MetaTroll.
\end{itemize}


\section{Related Work}

Our related work comes from three areas, troll detection, meta-learning and few-shot text classification. 

\subsection{Troll detection}
Early studies of troll detection focus on extracting hand-engineered features from the textual contents of user posts for troll detection~\citep{de2013filtering,dlala2014trolls,seah2015troll,dlala2014trolls,cheng2015antisocial,mihaylov2015exposing}. 
Signals such as writing style, sentiment as well as emotions have been explored~\citep{de2013filtering,seah2015troll}.
User online activities have also been used to detect trolls~\citep{dlala2014trolls,cheng2015antisocial}.
\citet{cheng2015antisocial} presents a study on anti-social behavior in online discussion communities, focusing on users that will eventually be banned.

More recently, approaches that combine user posts and online activities for troll detection have emerged~\citep{atanasov2019predicting, addawood2019linguistic, im2020still, shafiei2022detection, stewart2018modification, badawy2019falls, dutt2018senator}.
\citet{addawood2019linguistic} identify 49 linguistic markers of deception and measure their use by troll accounts. They show that such deceptive language cues can help to accurately identify trolls.  \citet{im2020still} propose a detection approach that relies on users' metadata, activity (e.g.\ number of shared links, retweets, mentions, etc.), and linguistic features to identify active trolls on Twitter.
\citet{shafiei2022detection} show that Russian trolls aim to hijack the political conversation to create distrust among different groups in the community.
\citet{stewart2018modification} similarly demonstrate how trolls act to accentuate disagreement and sow division along divergent frames, and this is further validated by \citet{dutt2018senator} in relation to Russian ads on Facebook. 
In \citet{badawy2019falls}, the authors study the effects of manipulation campaigns by analysing the accounts that endorsed trolls' activity on Twitter.
They find that conservative-leaning users re-shared troll content 30 times more than liberal ones. 
\citet{zannettou2019disinformation} compare troll behavior with other (random) Twitter accounts by recognising the differences in the content they spread, the evolution of their accounts, and the strategies they adopted to increase their impact.
\citet{atanasov2019predicting} leverage both social structure and textual contents to learn user representations via graph embedding.
We focus on how to detect trolls in emergent social activities with limited labelled data.

\subsection{Meta-learning}
Meta-learning  aims to extract some transferable knowledge from a set of {tasks} to quickly adapt to a new task.
These approaches can be divided into three categories: metric-based, optimisation-based, and model-based.
Metric-based meta-learning approaches~\cite{snell2017prototypical, vinyals2016matching} aim to learn an embedding function to encode the input and a metric function to learn the distance (e.g. cosine distance and euclidean distance) between the query data and support data.
Optimisation-based approaches~\citep{garnelo2018conditional,finn2017model} are designed to learn good parameter initialisation that can quickly adapt to new tasks within a few gradient descent steps.
Model-based approaches~\citep{santoro2016meta,munkhdalai2017meta} use neural networks to embed task information and predict test examples conditioned on the task information.
Our approach combines both optimisation and model-based ideas in that we adopt \cite{finn2017model} to update model parameters and a novel architecture that involves adapters and adaptive classification layers to learn task information.

\subsection{Few-shot text classification}
Text classification has shifted from task-specific training to pre-trained language models (such as BERT) followed by task-specific fine-tuning~\cite{devlin2018bert,pfeiffer2020adapterhub}. 
Recently, the language model GPT-3~\cite{brown2020language} has shown strong few-shot performance for many natural language processing tasks.
Few-shot text classification refer to a text classification setting where there are novel unseen tasks (domains) 
with only a few labelled examples for training a classification model, and 
meta-learning have been shown to produce strong performance 
~\citep{geng2019induction, gao2019hybrid, bao2020few,yu2018diverse}. 
\citet{bao2020few} introduce distributional signatures, such as word frequency and information entropy, into a meta-learning framework.
\citet{gao2019hybrid} combine attention mechanism with prototypical network to solve noisy few-shot relation classification task. 
\citet{geng2019induction} leverage the dynamic routing algorithm in meta-learning to solve sentiment and intent text classification tasks for English and Chinese datasets.
\citet{yu2018diverse} propose an adaptive metric-based method that can determine the best weighted combination automatically.
In our problem setting, our focus is on adapting the classifier to a new task (i.e.\ campaign) under the meta-learning framework.
\begin{figure*}[t]
\centering
  \includegraphics[width=0.78\textwidth]{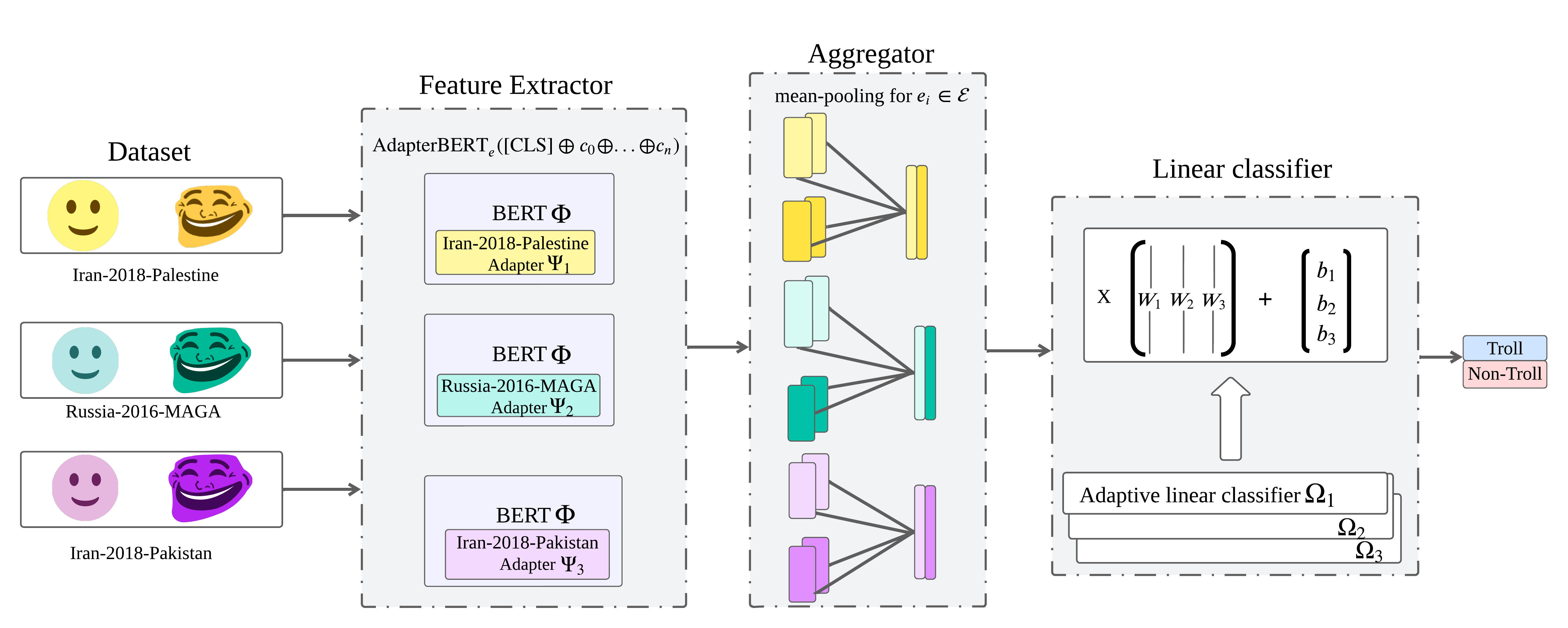}
  \caption{Overall architecture of MetaTroll.}
  \label{fig:model_architecture}
  \Description[Overall architecture of MetaTroll]{Overall architecture of MetaTroll}
\end{figure*}

\section{Problem Statement}
\label{sec:problem-statement}

We first explain some preliminaries of meta-learning, and contextualise it under our problem setting, before describing our model in \secref{approach}.
Let $\mathcal{E} = \{e_0, e_1, e_2,...,e_l\} $ be a set of campaigns. These campaigns are split into two partitions: 
$\mathcal{E}_{train}$ for meta-training and $\mathcal{E}_{test}$ for meta-testing (in our case, we have 6 campaigns for $\mathcal{E}_{train}$, and 4 for $\mathcal{E}_{test}$; see Table~\ref{tab:data}). 
For each  campaign $e$, we have labeled users $\mathcal{U}$, where each user $\mathbf{u}$ consists of a list of text posts and images, defined as $\mathbf{u} = \{(c_0, m_0, p_0), ..., (c_n, m_n, p_n)\}$, where $c$ refers to the textual content of the post,  $m$ a set of images, and $p$ the timestamp of the post.
Each user $\mathbf{u}$ is associated with a ground-truth label $y \in Y$, where $Y$ represents the label set (troll or non-troll; binary). 
Our goal is to leverage the knowledge learned from past (meta-train) campaigns ($\mathcal{E}_{train}$) to adapt quickly to new (meta-test) campaigns ($\mathcal{E}_{test}$)  via few-shot learning (e.g.\ 5 or 10 users for the new campaign).

In our problem setting, the notion of a {task} (from meta-learning) is the \textit{binary troll detection task for a particular campaign}.
Each task $\mathcal{T}_i = \left ( \mathcal{D}^s, \mathcal{D}^q \right )$ includes a support set $\mathcal{D}^s$ with $S$ data points and a query set $\mathcal{D}^q$ with $Q$ data points, where $ \mathcal{D}^s = \left\{\mathbf{u}_{i}^{s}, y_{i}^{s}\right\}_{i=1}^{S} , \mathcal{D}^q = \left\{\mathbf{u}_{i}^{q}, y_{i}^{q}\right\}_{i=1}^{Q}$, and $\mathbf{u}_{i}$ is a user with a list of posts and images and $y_{i} \in \left \{ 0,1 \right \}$ corresponding to troll/non-troll labels (as defined above).
During meta-training, we  iterate through different campaign in $\mathcal{E}_{train}$ and sample support and query instances to train the model. Once that's done, to adapt to a new campaigns in $\mathcal{E}_{test}$, we follow a similar training process but we update the model parameters using \textit{only} the support instances from the new campaign, and reserve the query set for evaluation (i.e.\ the query set is not used for model update).

\begin{algorithm*}
	\caption{MetaTroll}
	
	\begin{flushleft}
	\hspace*{\algorithmicindent} \textbf{Input/hyper-parameters:} meta-train  campaigns $\mathcal{E}_{train}$, meta-test  campaigns $\mathcal{E}_{test}$, learning rate $\beta$, $\gamma$ and $\delta$, model $M$ \\
	\hspace*{\algorithmicindent} \textbf{Parameters:}
	campaign-general adapter $\Psi$, campaign-specific adapter  $\Psi_e$, BERT $\Phi$, linear classifier $\Omega$, learning rate $\alpha$\\
    \end{flushleft}
    
	\begin{algorithmic}[1]
	\State {\#\#\# Meta-training \#\#\#}
	\State Fine-tune base $M$ on $\mathcal{E}_{train}$ and provide initial model parameters $\Phi$ \phantom{xx} {\#Stage 1}
	\While{$\textit{not done}$} \phantom{xx} {\#Stage 2}
		\For {$e \in \mathcal{E}_{train}$}
			\State Sample batch of tasks $\mathcal{T}_i \in \mathcal{T}$
			\For {each $\mathcal{T}_i$}
				\State Sample $S$ data-points to form $\mathcal{D}^{s} =\left\{\left(\mathbf{u}_i, y_i\right)\right\}_{i=1}^S \in \mathcal{T}_i$ as support set
				\State Sample $Q$ data-points to form $\mathcal{D}^{q} =\left\{\left(\mathbf{u}_i, y_i\right)\right\}_{i=1}^Q \in \mathcal{T}_i$ as query set for meta-update
				\State Compute $\nabla_{\Psi} \mathcal{L}\left(M_{\Phi,\Psi}\right)$ on $D^{s} $
				\State Compute adapted parameters for adapter: $\Psi^{\prime}=\Psi - \alpha \nabla_{\Psi} \mathcal{L}_{\mathcal{T}_i}\left(M_{\Phi,\Psi} \right)$
				\State Compute $\mathcal{L}_{\mathcal{T}_i}\left(M_{\Phi,\Psi^{\prime}}\right)$ on $D^{q}$
			\EndFor
		
		\State Update adapter parameters: $\Psi \leftarrow \Psi -\beta \nabla_{\Psi} \mathbb{E}_{\mathcal{T}_i \sim \mathcal{T}}\left[\mathcal{L}_{\mathcal{T}_i}\left(M_{\Phi,\Psi^{\prime}}, \mathcal{D}^{q}\right)\right]$
		\State Update inner loop learning rate: $\alpha \leftarrow \Psi-\beta \nabla_{\alpha} \mathbb{E}_{\mathcal{T}_i \sim \mathcal{T}}\left[\mathcal{L}_{\mathcal{T}_i}\left(M_{\Phi,\Psi^{\prime}}, \mathcal{D}^{q}\right)\right]$
		\EndFor
	\EndWhile
	\While{$\textit{not done}$} \phantom{xx}  {\#Stage 3}
	\For {$e \in \mathcal{E}_{train}$}
		\State Sample batch of tasks $\mathcal{T}_i \in \mathcal{T}$		
		\For {each $\mathcal{T}_i $}
			\State Sample $S$ data-points to form $\mathcal{D}^{s} =\left\{\left(\mathbf{u}_i, y_i\right)\right\}_{i=1}^S \in \mathcal{T}_i$ as support set
			\State Sample $Q$ data-points to form $\mathcal{D}^{q} =\left\{\left(\mathbf{u}_i, y_i\right)\right\}_{i=1}^Q \in \mathcal{T}_i$ as query set for meta-update
			\State Compute mean troll and non-troll representations based on support set 
			\State Compute $\nabla_{\Psi_e,\Omega_e} \mathcal{L}\left(M_{\Phi, \Psi_e,\Omega_e}\right)$ on $D^{s} $
			\State Compute adapted parameters for adapter: $\Psi^{\prime}_e =\Psi_e - \gamma \nabla_{\Psi_e} \mathcal{L}_{\mathcal{T}_i}\left(M_{\Phi, \Psi_e,\Omega_e} \right)$ 
			\State Compute adapted parameters for classifier: $\Omega^{\prime}_e =\Omega_e - \gamma \nabla_{\Omega_e} \mathcal{L}_{\mathcal{T}_i}\left(M_{\Phi, \Psi_e,\Omega_e} \right)$ 
			\State Compute $\mathcal{L}_{\mathcal{T}_i}\left(M_{\Phi, \Psi^{\prime}_e,\Omega^{\prime}_e}\right)$ on $D^{q}$
		\EndFor
	
        \State Update campaign-specific adapter parameters: $\Psi_e \leftarrow \Psi_e - \delta \nabla_{\Psi_e} \mathbb{E}_{\mathcal{T}_i \sim \mathcal{T}}[\mathcal{L}_{\mathcal{T}_i}(M_{\Phi,\Psi^{\prime}_e, \Omega^{\prime}_e}, \mathcal{D}^{q})]$
		\State Update classifier parameters: $\Omega_e \leftarrow \Omega_e - \delta \nabla_{\Omega_e} \mathbb{E}_{\mathcal{T}_i \sim \mathcal{T}}[\mathcal{L}_{\mathcal{T}_i}(M_{\Phi,\Psi^{\prime}_e, \Omega^{\prime}_e}, \mathcal{D}^{q})]$
	\EndFor
	\EndWhile
	\State {\#\#\# Meta-testing \#\#\#}
	\State Perform few-shot learning on meta-test campaigns $\mathcal{E}_{test}$ using meta-learned parameters $\left ( \Phi, \Psi_e, \Omega_e  \right )$
	\end{algorithmic}
	\label{alg:metaTroll_alg} 
\end{algorithm*} 

\section{Approach}
\label{sec:approach}



We present the overall architecture of MetaTroll in Figure~\ref{fig:model_architecture}. MetaTroll has two modules, a BERT-based ~\cite{devlin2018bert} feature extractor and an adaptive linear classifier. 

We first explain at a high level how we train MetaTroll in three stages for the feature extractor and linear classifier (Algorithm~\ref{alg:metaTroll_alg}).
In the first stage, we fine-tune an off-the-shelf BERT and update all its parameters ($\Phi$) 
using \textit{all} training samples from the meta-train campaigns $\mathcal{E}_{train}$ (line 1 in Algorithm~\ref{alg:metaTroll_alg}). At this stage we do not introduce adapters \cite{houlsby2019parameter} to BERT and it is optimised to do binary classification of trolls and non-trolls. The idea of this step is to train a feature extractor for the troll detection task (in other words, this step is standard fine-tuning to adapt BERT for binary troll detection).

In the second stage, we introduce adapters \cite{houlsby2019parameter} to BERT and train the adapter parameters ($\Psi$) using model agnostic
meta-learning (MAML: \citet{finn2017model}; line 3--16 in Algorithm~\ref{alg:metaTroll_alg}). Note that the adapter is \textit{shared} across all campaigns (i.e.\ the adapter is \textit{not} campaign-specific), and the idea of this stage is to learn a good initialisation for the adapter to do general troll detection; in the next stage, the learned adapter parameters will be used to initalise \textit{campaign-specific} adapters.
During this stage, only the adapter parameters ($\Psi$) are updated while the BERT parameters ($\Phi$) are frozen. 

In the third and final stage, we introduce campaign-specific adapters 
and adaptive linear classifiers to MetaTroll, creating the full model. The campaign-specific adapters are initialised using the campaign-general adapter from stage 2. The idea of using campaign-specific adapters is to address catastrophic forgetting when MetaTroll is continuously updated for new campaigns that emerge over time in an application setting: the campaign-specific adapter solves the `forgetting' problem because the knowledge of detecting older/past campaigns are stored in their adapters which will not be overwritten as MetaTroll is continually updated. As for the adaptive linear classifiers, they are also campaign-specific and designed to encourage MetaTroll to learn campaign representations that are distinct for different campaigns.\footnote{The adaptive linear classifier parameters are initialised randomly for a new campaign.}
We update the campaign-specific adapter  ($\Psi_{e}$) and classifier parameters ($\Omega_e$) via meta-learning, similar to stage 2 training (line 17--32 in  Algorithm~\ref{alg:metaTroll_alg}), noting that the BERT parameters ($\Phi$) are frozen in this stage.

After MetaTroll is trained (over the 3 stages), to adapt it to trolls of a new campaign at test time, we follow the third stage training to learn campaign-specific adapter ($\Psi_{e}$) and classifier ($\Omega_e$) parameters for the new campaign. Once adapted, MetaTroll can be used to classify users from this new campaign. We next describe the training stages and test inference in detail.



%

\subsection{Stage One Training}
\label{sec:stage-one}

In the first stage, MetaTroll is a standard BERT fine-tuned with standard cross entropy loss to do binary classification of trolls vs. non-trolls (ignoring the campaigns). Given a user $\mathbf{u}$ with posts \{$c_0,...,c_n$\}: 
\begin{align}
    v &= \text{BERT}([\text{CLS}] \oplus c_0 \oplus ... \oplus c_n) \label{eqn:bert} \\ 
    \hat{y} &= \text{softmax} (W v + b)  \label{eqn:softmax}
\end{align}
where $v$ is the contextual embedding of $[$CLS$]$ and $\oplus$ is the concatenation operation. 



\begin{figure}[tbh]
\centering
  \includegraphics[width=0.55\linewidth]{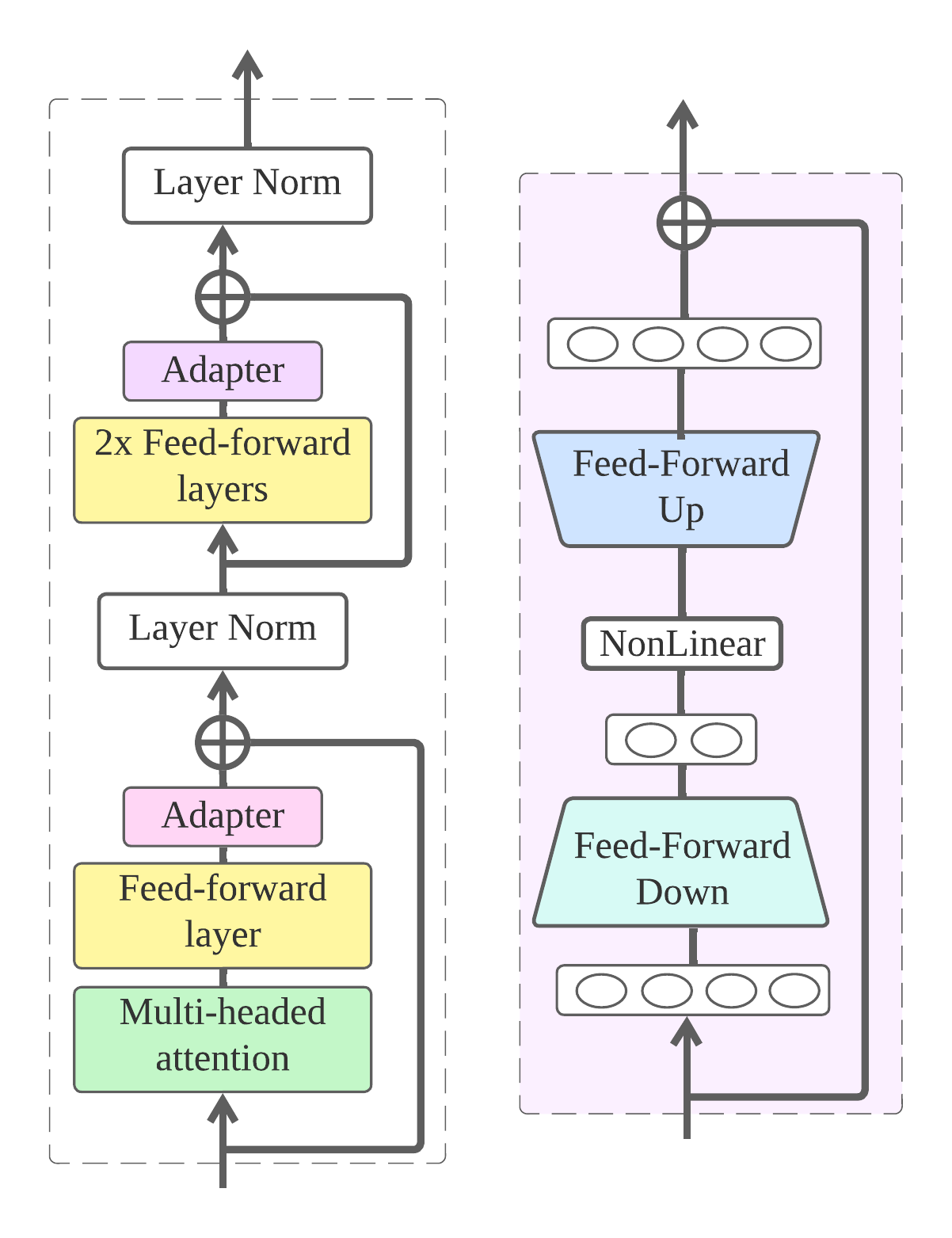}
  \caption{Overall architecture of Adapter-BERT. The left figure illustrates how two adapter modules are added to a transformer layer; right shows the components in the adapter.}
  \label{fig:adapter_architecture}
  \Description[Overall architecture of Adapter-BERT]{Overall architecture of Adapter-BERT. The left figure illustrates how two adapter modules are added to a transformer layer; right shows the components in the adapter.}
  \vspace{-1\baselineskip}
\end{figure}

\subsection{Stage Two Training}
\label{sec:stage-two}

In the second stage, we add adapters to BERT (see Figure~\ref{fig:adapter_architecture}).
Following~\cite{houlsby2019parameter}, we insert two adapter modules containing bottleneck layers into each transformer layer of BERT.
Note that the adapter is \textit{shared} across all campaigns here, as the goal in this stage is to learn good set of initial adapter parameter values that can be used to initialise campaign-specific adapters in the next stage.

Correspondingly, Equation \ref{eqn:bert} is now modified to:
\begin{align}
    v = \text{AdapterBERT}([\text{CLS}] \oplus c_0 \oplus ... \oplus c_n) \label{eqn:adapterbert}
\end{align}

We meta-train the adapter parameters using MAML (line 3--16 in Algorithm~\ref{alg:metaTroll_alg}).
We first sample a task $\mathcal{T}_{i}$ for campaign $e$ from  $\mathcal{E}_{train}$ to create the support $D^{s}$ and query set  $D^q$.
Denoting M as the model, $\Phi$ the BERT parameters and $\Psi$ the adapter parameters, we next compute the inner loop to update $\Psi$ as follows \cite{li2017meta}:
\begin{align*}
    \mathcal{L}_{\mathcal{T}_{i}}(M_{\Phi,\Psi}, \mathcal{D}^{s}) &= \frac{1}{|\mathcal{D}^{s}|} \sum_{\mathbf{u}_i \in \mathcal{D}^{s}}-\log p\left({y}_i|\mathbf{u}_i ; \Phi, \Psi \right) \\
       \Psi^{\prime} &= \Psi - \alpha \nabla_{\Psi} \mathcal{L}\left(M_{\Phi,\Psi}, \mathcal{D}^{s}\right) \label{eqn:inner}
\end{align*}
where $\alpha$ is the learning rate (computed next in the outer loop)
and $\mathcal{L}_{\mathcal{T}_{i}}(M_{\Phi,\Psi}, \mathcal{D}^{s})$ is the cross entropy loss over the support set. Next we compute the cross-entropy loss based on the query set, using the updated parameters:
\begin{equation*}
        \mathcal{L}_{\mathcal{T}_{i}}(M_{\Phi,\Psi^{\prime}}, \mathcal{D}^{q}) = \frac{1}{|\mathcal{D}^{q}|} \sum_{\mathbf{u}_i \in \mathcal{D}^{q}}-\log p\left({y}_i|\mathbf{u}_i ; \Phi, \Psi^\prime \right)
\end{equation*}

This inner loop is carried out for multiple steps of gradient descent (using several tasks from the same campaign). Note that in this stage  $\Phi $ (BERT parameters) is frozen. Once that's done, we update the adapter parameters $\Psi$ and inner loop learning rate $\alpha$:\footnote{In our implementation, $\alpha$ is layer-specific, i.e.\ we have a separate learning rate for each adapter in different layers.}
\begin{equation*}
    \begin{aligned}
        \Psi \leftarrow \Psi-\beta \nabla_{{\Psi}} \mathbb{E}_{\mathcal{T}_i \sim \mathcal{T}}\left[\mathcal{L}_{\mathcal{T}_i}\left(M_{\Phi,\Psi^{\prime}}, \mathcal{D}^{q}\right)\right]
        \\
        \alpha \leftarrow \Psi-\beta \nabla_{\alpha} \mathbb{E}_{\mathcal{T}_i \sim \mathcal{T}}\left[\mathcal{L}_{\mathcal{T}_i}\left(M_{\Phi,\Psi^{\prime}}, \mathcal{D}^{q}\right)\right]
    \end{aligned}
\end{equation*}
where $\beta $ is the learning rate for the outer loop (set to $1e^{-5}$ in our experiments).

\subsection{Stage Three Training}
\label{sec:stage-three}

In this stage, we introduce campaign-specific adapters for each campaign, and they are all initialised using the campaign-general adapter learned from the previous stage. Formally, Equation \ref{eqn:adapterbert} is now updated to:
\begin{align*}
    v = \text{AdapterBERT}_e([\text{CLS}] \oplus c_0 \oplus ... \oplus c_n)
\end{align*}
where $e$ is the campaign.

We introduce campaign-specific adapters to our model for two reasons: (1) they  are  more efficient to train  and less vulnerable to overfitting (which is important in a few-shot learning setting), since they contain only a small number of parameters compared to the alternative where we have one BERT model for every campaign); and (2)  they alleviate catastrophic forgetting in a continual learning setting, as each campaign has its own adapter.


Inspired by \citet{requeima2019fast}, we next introduce an adaptive linear classifier network that replaces $W$ and $b$ in the linear layer used for classification (Equation \ref{eqn:softmax}). Intuitively, this adaptive classifier works by first computing the aggregate troll and non-troll representation for each campaign, and then learn campaign-specific projections to classify between trolls vs.\ non-trolls.
Let $\mathcal{D}_{0}^{s}$ and $\mathcal{D}_1^s$ denote the support set where the labels are trolls ($y=0$) and non-trolls ($y=1$) respectively, we compute $W_e$, $b_e$ and $\hat{y}$ for troll campaign $e$ as follows:
\begin{align*}
	W_{e}^{0} &= \frac{1}{\left|\mathcal{D}_{0}^s\right|} \sum_{v_ \in \mathcal{D}_{0}^s}( v ) \nonumber \quad 
	b_e^{0} = \frac{1}{\left|\mathcal{D}_{0}^s \right|} \sum_{v_ \in \mathcal{D}_{0}^s}( v ) \nonumber \\
	W_{e}^{1} &= \frac{1}{\left|\mathcal{D}_{1}^s\right|} \sum_{v_ \in \mathcal{D}_{1}^s}( v ) \nonumber \quad 
	b_e^{1} = \frac{1}{\left|\mathcal{D}_{1}^s \right|} \sum_{v_ \in \mathcal{D}_{1}^s}( v ) \nonumber \\
	\hat{y} &= \text{softmax} (W_e v + b_e)  
\end{align*}
where $W^i_e$ denotes the $(i-1)^\text{th}$ column of $W_e$.
In other words, MetaTroll classifies a user based on whether its representation ($v$) is closer to the (average) troll or non-troll representation.

The campaign-specific adaptive linear classifier parameters $\Omega_e$ and the adapters parameters $\Psi_e$ are trained using MAML, just like stage 2 training (line 17--31 in  Algorithm~\ref{alg:metaTroll_alg}).

\subsection{Meta-testing}

After MetaTroll is trained, to adapt it to a new campaign $e$ , we follow the process of the third stage training. To simulate few-shot learning, we sample only a small number of instances for the support and query set (e.g.\ 5 each), and use \textit{only} the support set for updating the adapter  ($\Psi_e$)  and classifier parameters ($\Omega_e$) (line 25--26 in Algorithm~\ref{alg:metaTroll_alg}) and do not run the outer loop ((line 30--31). Here the query set is used only for computing performance (i.e.\ in line 27 we compute accuracy instead of loss over the query set).

\section{Experiments and Results}
\begin{table*}[t]
\caption{Statistics of English meta-train and meta-test data. The last row is non-troll users that are sampled by generating random user Twitter IDs.}
\vspace{-1\baselineskip}
\small
\begin{center}
\begin{tabular}{lccccr@{\;\;}c}
\toprule
\midrule
&Train/Test &Campaign &Event Time & Type & \#users  & Top-3 Hashtags\\ \midrule
&\multirow{12}{*}{Meta-Train}&\multirow{2}{*}{Iran-2018-Palestine}  &\multirow{2}{*}{Feb 2018 -- Aug 2018}  & Troll & 557  & \multirow{2}{*}{\shortstack{\#realiran \#SavePalestine \\ \#InternationalQudsDay2018 }} \\
& & & & Non-troll  & 1,063 \\ \cline{3-7}
& &\multirow{2}{*}{Russia-2016-MAGA}  &\multirow{2}{*}{Aug 2015 -- Feb 2016 }  & Troll & 247 
&\multirow{2}{*}{\shortstack{\#MAGA \#QAnon \\ \#ReleaseTheMemo}} \\
& & & & Non-troll & 663   \\ \cline{3-7}
&&\multirow{2}{*}{Iran-2018-Pakistan} &\multirow{2}{*}{May 2018 -- Nov 2018}  & Troll  & 2,267  &\multirow{2}{*}{\shortstack{\#pakonlinenews \\ \#SachTimes \#DeleteIsrael}} \\
& & & & Non-troll & 2,500  \\ \cline{3-7}
& &\multirow{2}{*}{Venezuela-2018-Trump} &\multirow{2}{*}{Jun 2018 -- Dec 2018} & Troll  & 1,330  &\multirow{2}{*}{\#TrumpTrain \#MAGA \#RT}\\
& & & & Non-troll   & 1,500   \\ \cline{3-7}
& &\multirow{2}{*}{Nigeria-2019-Racism} &\multirow{2}{*}{Aug 2019 -- Feb 2020}  & Troll  & 1,120 
&\multirow{2}{*}{\shortstack{\#racism \\ \#BlackLivesMatter \#PoliceBrutality}} \\
& & & & Non-troll  & 1,200 \\ \cline{3-7}
& &\multirow{2}{*}{Iran-2020-BLM}  &\multirow{2}{*}{Jul 2020 -- Jan 2021} & Troll & 205 
&\multirow{2}{*}{\shortstack{\#black\_lives\_matter \\ \#Oscars \#EEUU}} \\
& & & & Non-troll &  212   \\  \midrule
&\multirow{8}{*}{Meta-Test} &\multirow{2}{*}{GRU-2020-NATO}   &\multirow{2}{*}{Jun 2020 -- Dec 2020} & Troll & 35 &
\multirow{2}{*}{\shortstack{\#Syria \#Idib \\ \#StopTerrorismInSyria}} \\
& & & & Non-troll &  135  \\ \cline{3-7}
& &\multirow{2}{*}{IRA-2020-Russia}  &\multirow{2}{*}{Jun 2020 -- Dec 2020} & Troll  & 20 & \multirow{2}{*}{\#valdaiclub \#Russia \#Ukraine} \\
& & & & Non-troll  & 81  \\ \cline{3-7}
& &\multirow{2}{*}{Uganda-2021-NRM}   &\multirow{2}{*}{Jul 2020 -- Jan 2021} & Troll & 334 & \multirow{2}{*}{\shortstack{\#SecuringYourFuture \\ \#M7UGsChoice \#StopHooligansim}} \\
& & & & Non-troll  & 542  \\ \cline{3-7}
& &\multirow{2}{*}{China-2021-Xinjiang}  &\multirow{2}{*}{Jul 2020 -- Jan 2021} & Troll & 3,440  &\multirow{2}{*}{\shortstack{\#Xinjiang \\ \#XinjiangOnline \#Urumqi}} \\
& & & & Non-troll &  2,345 \\ \midrule
& &\multirow{2}{*}{Random}  &\multirow{2}{*}{Varies} & \multirow{2}{*}{Non-troll} &\multirow{2}{*}{8,000} &\multirow{2}{*}{\shortstack{\#entertainment \\ \#SoundCloud \#Vegas}} \\
& & & & & \\
\midrule
\bottomrule
\end{tabular}
\end{center}
\label{tab:data}
\end{table*}

\subsection{Datasets and models}
\label{sec:datasets}

We use the information operations dataset published by Twitter for our experiments.\footnote{\url{https://transparency.twitter.com/en/reports/information-operations.html}}
This dataset contains different groups of users banned by Twitter since October 2018 for engaging in state-sponsored information operations, and each group represents a \textit{campaign} in our work.
For example, the ``Iran-2018-Palestine'' campaign refers to trolls sponsored by Iran for an information campaign targeting Palestine in 2018.\footnote{\url{https://blog.twitter.com/en_us/topics/company/2018/enabling-further-research-of-information-operations-on-twitter}} To clarify, these campaigns are defined by Twitter when they release the data, and each campaign is associated with a blogpost that explains the information operation.

For each campaign, we filter users and keep only those who have posted a tweet within the 6-month event period (``Event Time'' in \tabref{data}) to remove users who are inactive during the event.\footnote{The event period is determined as follows: (1) the end date is the last post in the campaign; and the (2) start date is 6 months from the end date.} For each user, we also filter their tweets to keep only their most recent 20 posts that have a timestamp within the event period.

To create the non-troll users, we combine two sources: (1) ``Random'', random accounts that are sampled by generating random numeric user IDs and validating their existence following \cite{alizadeh2020content}; and (2) ``Hashtag'', users whose posts contain popular hashtags used by a campaign, where popular hashtags are defined as hashtags that collectively cover 75\% of trolls' posts. The reason why we have two types of non-troll users is that if we only sample random users as non-trolls, the post content of non-trolls would be \textit{topically} very different to that of the trolls, and the detection task would degenerate into a topic detection task. The non-troll users sampled using the ``Hashtag'' approach is designed to circumvent this and makes the detection task more challenging.

Table~\ref{tab:data} present some statistics for trolls and non-trolls in different campaigns, where 6 are used for meta-training and 4 for meta-testing.
Note that the non-trolls of a particular campaign are users sampled with the ``Hashtag'' approach, and the last row corresponds to non-troll users sampled using the ``Random'' approach.
For these campaigns, at least 80\% of the trolls' posts are in English, and so they are used for the monolingual (English) experiments in the paper.\footnote{For non-troll users, we only keep English tweets (based on the predicted language given in the metadata).}


The trolls and non-trolls in Table~\ref{tab:data} represent the pool of users which we draw from to construct the final training/testing data. In all of our experiments, we keep the ratio of trolls to non-trolls to 50/50 through sampling,
and when sampling for non-troll users, the ratio from the two sources (``Random'' and ``Hashtag'') is also 50/50.\footnote{We also only sample ``Random'' users whose most recent post is in the event time.} As an example,  ``Uganda-2021-NRM'' has 334 troll users. We therefore sample 334 non-troll users, where 167 are from ``Random'' and another 167 from ``Hashtag''.


We compare our MetaTroll model against the following baselines including state-of-the-art meta-learning methods and  few-shot text classification models:
\begin{itemize}
\setlength\itemsep{0em}
\item \textbf{BERT}~\cite{devlin2018bert}\footnote{\url{https://huggingface.co/docs/transformers/model_doc/bert}}: BERT fine-tuned using the support set of meta-test data.
\item \textbf{KNN}~\cite{khandelwalgeneralization}: $K$-nearest neighbour classifier with off-the-shelf BERT as the feature extractor.\footnote{$K$ is selected from [5,10].}
\item \textbf{AdBERT}~\cite{pfeiffer2020adapterhub}\footnote{\url{https://adapterhub.ml/}}: BERT that fine-tunes an adapter for each campaign.
\item \textbf{GPT3}~\cite{brown2020language}\footnote{\url{https://gpt3demo.com/apps/openai-gpt-3-playground}}: a very large pretrained model adapted to our tasks using prompt-based learning \cite{liu2021pre}.\footnote{We use ``text-davinci-002'' in our experiments. The prompts are a small set of instances (e.g.\ 5) with their labels added to the beginning of the input.}
\item \textbf{MAML}~\cite{finn2017model}\footnote{\url{https://github.com/tristandeleu/pytorch-meta/tree/master/examples/maml}}: BERT trained with the MAML algorithm for few-shot learning.
\item \textbf{CNP}~\cite{garnelo2018conditional}\footnote{\url{https://github.com/deepmind/neural-processes}}: a model-based meta-learning framework that consists of a shared encoder, aggregator and decoder.
\item \textbf{ProtoNet}~\cite{snell2017prototypical}\footnote{\url{https://github.com/orobix/Prototypical-Networks-for-Few-shot-Learning-PyTorch}}: a deep metric-based approach using sample average as class prototypes and the distance is calculated based on euclidean-distance. 
\item \textbf{Induct}~\cite{geng2019induction}: a few-shot classification model that uses a dynamic routing algorithm to learn a class-wise representation.
\item \textbf{HATT}~\cite{gao2019hybrid}\footnote{\url{https://github.com/thunlp/HATT-Proto}}: A classification model of metric-based meta-learning framework together with attention mechanism.
\item \textbf{DS}~\cite{bao2020few}\footnote{\url{https://github.com/YujiaBao/Distributional-Signatures}}: A few-shot text classification model that uses distributional signatures such as word frequency and information entropy for training.
\end{itemize}

\begin{table*}[t]
\caption{\label{tab:english_result} Troll detection accuracy in 5-shot and 10-shot settings.}
\vspace{-1\baselineskip}
\small
\begin{center}
\begin{tabular}{lc@{\;\;}cc@{\;\;}cc@{\;\;}cc@{\;\;}c@{\;\;}cc@{\;\;}c}
\toprule
\toprule
      & \multicolumn{2}{c}{GRU-2020-NATO} & \multicolumn{2}{c}{IRA-2020-Russia} &\multicolumn{2}{c}{Uganda-2021-NRM} &\multicolumn{2}{c}{China-2021-Xinjiang} \\ \midrule
Model & 5-shot   & 10-shot  & 5-shot   & 10-shot  &  5-shot   & 10-shot  &  5-shot   & 10-shot \\ \toprule
BERT~\cite{devlin2018bert} &50.41 &51.39$\uparrow_{0.98}$ &51.70 &52.21$\uparrow_{0.51}$ &55.96 &56.53$\uparrow_{0.39}$ &60.36 &61.68$\uparrow_{1.32}$ \\
KNN~\cite{khandelwalgeneralization}   &46.10  &48.91$\uparrow_{2.81}$ &42.52  &44.28$\uparrow_{1.76}$ &46.74 &48.49$\uparrow_{1.75}$ &50.21 &51.85$\uparrow_{1.37}$ \\ 
AdBERT~\cite{pfeiffer2020adapterhub} &65.05 &66.73$\uparrow_{1.68}$ &55.16 &57.39$\uparrow_{2.23}$ &52.24 &56.27$\uparrow_{4.03}$ &65.49 &67.38$\uparrow_{1.89}$ \\ 
GPT3~\cite{brown2020language} &54.25 &71.39$\uparrow_{17.14}$ &61.87 &75.96$\uparrow_{14.09}$ &52.49 &69.66$\uparrow_{17.17}$ &70.34 &80.49$\uparrow_{10.15}$ \\
\midrule 
MAML~\cite{finn2017model} &70.13 &72.43$\uparrow_{2.30}$ &70.11 &71.82$\uparrow_{1.71}$ &61.11 &65.74$\uparrow_{4.63}$ &70.99 &73.87$\uparrow_{2.85}$ \\
CNP~\cite{garnelo2018conditional} &69.88 &71.26$\uparrow_{1.38}$ &69.47 &72.70$\uparrow_{3.23}$ &64.48 &66.95$\uparrow_{2.47}$ &67.05 &69.95$\uparrow_{2.90}$ \\
ProtoNet~\cite{snell2017prototypical} &60.43 &62.87$\uparrow_{2.44}$ &67.87 &70.05$\uparrow_{2.18}$ &64.43 &67.79$\uparrow_{3.36}$ &74.56 &76.65$\uparrow_{2.09}$\\
\midrule
Induct~\cite{geng2019induction} &71.11 &72.26$\uparrow_{1.15}$ &\textbf{76.49} &77.24$\uparrow_{0.75}$ &71.55 &73.50$\uparrow_{1.95}$ &75.13 &76.70$\uparrow_{1.57}$ \\
HATT~\cite{gao2019hybrid} &63.02 &65.56$\uparrow_{2.54}$ &75.82 &77.50$\uparrow_{1.68}$ &60.45 &63.87$\uparrow_{3.42}$ &78.22 &80.03$\uparrow_{1.81}$ \\
DS~\cite{bao2020few} &62.69 &64.62$\uparrow_{1.93}$ &66.05 &68.92$\uparrow_{2.87}$ &62.43 &64.50$\uparrow_{2.07}$ &71.74 &73.43$\uparrow_{1.69}$ \\
\midrule
MetaTroll &\textbf{72.74} &\textbf{75.13} $\uparrow_{2.39}$ &76.25 &\textbf{77.99}$\uparrow_{1.74}$ &\textbf{75.05}&\textbf{78.25}$\uparrow_{3.20}$ &\textbf{81.37} &\textbf{84.50}$\uparrow_{3.13}$ \\ \bottomrule
\bottomrule
\end{tabular}
\end{center}
\vspace{-1\baselineskip}
\end{table*}

\subsection{Results}

We first present English troll detection performance. In Sections \ref{sec:multilingual} and \ref{sec:multimodal} we extend MetaTroll to work with non-English data and images. As we focus on English here, non-English posts are discarded (although this only reduces the data size by a negligible amount, as most data is in English, as explained in \secref{datasets}).
All reported figures are an average accuracy performance over 5 runs with different random seeds.\footnote{Noting that the performance for one run is an average performance over the query sets from multiple tasks.
}

Table~\ref{tab:english_result} presents 5-shot and 10-shot results for the 4 meta-test campaigns.\footnote{5-shot means only 5 labelled instances for each class are given in the support set.}
MetaTroll is the best model, achieving an average of 76.35\% accuracy over all campaigns. That said, some of the few-shot text classifiers (Induct e.g.73.57\%) are not far behind.
Most models only benefit marginally by seeing 5 more examples going from 5- to 10-shot, with the exception of GPT3 where we see a substantial performance boost (average 14.64\% gain).

\begin{table}[t]
\small
\caption{\label{tab:seq_result} Catastrophic forgetting results under continual learning. ``G'' $=$ GRU-2020-NATO, ``I'' $=$ IRA-2020-Russia, ``U'' $=$ Uganda-2021-NRM, ``C'' $=$ China-2021-Xinjiang.}  
\vspace{-0.5\baselineskip}
\begin{center}
 \begin{tabular}{lccc@{\;\;}cc@{\;\;}c@{\;\;}c}
\toprule
\toprule
\multirow{2}{*}{Model} & & \multicolumn{1}{c}{G$\rightarrow$I} & \multicolumn{2}{c}{G$\rightarrow$I$\rightarrow$U} & \multicolumn{3}{c}{G$\rightarrow$I $\rightarrow$U$\rightarrow$C} \\
\cmidrule(l{.75em}r{.75em}){3-3}\cmidrule(l{.75em}r{.75em}){4-5}\cmidrule(l{.75em}r{.75em}){6-8}
   &G &G  &G & I  &G & I & U \\ \toprule
BERT~\cite{devlin2018bert}  &50.41 &45.16 &48.02 &44.80 &42.85 &43.25 & 50.00 \\
KNN~\cite{khandelwalgeneralization}  &46.10 &45.25  &43.12 &49.99 &43.01 &45.50 &45.33 \\ 
AdBERT~\cite{pfeiffer2020adapterhub} &65.05 &62.41 &54.59 &52.56 &50.54 &53.74 &49.95 \\
GPT3~\cite{brown2020language} &54.25 &51.92 &50.33 &58.88 &48.93 &56.87 &50.06 \\
\midrule
MAML~\cite{finn2017model} &70.13 &68.32 &65.15 &56.55 &57.62 &50.04 &54.87 \\
CNP~\cite{garnelo2018conditional} &69.88 &71.55 &68.72 &56.75 &60.57 &56.95 &63.21 \\
ProtoNet~\cite{snell2017prototypical} &60.43 &56.56 &61.05 &61.99 &53.47 &52.83 &60.05 \\
\midrule
Induct~\cite{geng2019induction} &71.11 &69.50 &62.82 &66.13 &58.43 &63.52 &64.83 \\  
HATT~\cite{gao2019hybrid} &63.02 &60.50 &55.62 &70.18 &50.25 &50.11 &55.43 \\
DS~\cite{bao2020few} &62.69 &60.05 &58.85 &61.50 &50.65 &60.12 &61.85 \\
\midrule
MetaTroll &72.74 &\textbf{73.45} &\textbf{72.18}  &\textbf{70.81} &\textbf{71.15} &\textbf{69.74} &\textbf{68.75} \\
\bottomrule
\bottomrule
\end{tabular}
\end{center}
\vspace{-1\baselineskip}
\end{table}

\begin{table}[t]
\small
\caption{\label{tab:category_acc} Campaign classification results under continual learning.} 
\vspace{-0.5\baselineskip}
\begin{center}
\begin{adjustbox}{max width=1.0\linewidth}
 \begin{tabular}{lcc@{\;\;}cc@{\;\;}c@{\;\;}c}
\toprule
\toprule
\multirow{2}{*}{Model} & \multicolumn{1}{c}{G$\rightarrow$I} & \multicolumn{2}{c}{G$\rightarrow$I$\rightarrow$U} & \multicolumn{3}{c}{G$\rightarrow$I $\rightarrow$U$\rightarrow$C} \\
\cmidrule(l{.75em}r{.75em}){2-2}\cmidrule(l{.75em}r{.75em}){3-4}\cmidrule(l{.75em}r{.75em}){5-7}
   &I  &I & U  &I & U & C \\ \toprule
GPT3~\cite{brown2020language} &74.86 &79.25 &82.86 &81.17 &79.38 &74.62 \\
\midrule
MetaTroll &85.57 &85.15 &87.50 &83.61 &86.92 &86.11 \\
\bottomrule
\bottomrule
\end{tabular}
\end{adjustbox}
\end{center}
\vspace{-1\baselineskip}
\end{table}

\subsection{Continual learning performance}

In continual learning~\cite{ring1998child}, new tasks appear over time, 
where the goal of learning is to adapt the model accordingly to the new tasks without forgetting the previous tasks.
This is a more realistic setting for a troll detection system, as it should continually adapt to new campaigns that appear over time.
But in this setting it will suffer from catastrophic forgetting \citep{yap2021addressing,Xu+:2020}, where after adapting to newer campaigns its performance to classify older campaigns will degrade.

To simulate this continual learning setting, we next evaluate the troll detection models on a \textit{past campaign} after  
it has been adapted for a number of campaigns in sequence. 
For example, a system is first adapted to GRU-2020-NATO (G), and then to IRA-2020-Russia (I), Uganda-2021-NRM (U) and China-2021-Xinjiang (C) in sequence (denoted as G$\rightarrow$I$\rightarrow$U$\rightarrow$C). 
We then test the system using trolls using the past campaigns, i.e.\ G, I and U.


One challenge with MetaTroll under this continual learning evaluation is that at test time it needs to know which adapter to use --- information that most other systems do not require as they don't have campaign-specific parameters (exception: AdBERT  and GPT-3\footnote{GPT3 technically does not have campaign-specific parameters, but it needs to be given campaign-specific prompts, and so requires the campaign label.}). We experiment a simple approach to solve this: have MetaTroll classify a user using all its adapters, and select the outcome that has the highest probability.\footnote{We do the same for AdBERT and GPT3.}

We present (5-shot) troll classification results under this continual learning setting in 
Table~\ref{tab:seq_result}.
AdBERT and GPT-3 suffers little catastrophic forgetting, as they have campaign-specific parameters or prompts (they are unaffected by continual learning as their base model is unchanged), although their performance is only marginally above random chance in the first place.
MetaTroll is the clear winner here, with only $<$5\% accuracy degradation over time.\footnote{Note that performance of these systems on older campaigns will still degrade slightly over more adaptations, as there are more campaign-specific adapters or prompts to select from.}
In contrast, the meta-learning methods and few-shot classifiers suffer from catastrophic forgetting and their performance on older campaigns drops substantially (e.g. MAML's performance for GRU-2020-NATO drops from 68.32 to 65.15 and 57.62 after several updates).

An unintended benefit of that MetaTroll classifies trolls using multiple adapters is that it is effectively doing multi-class rather than binary classification (with over 85\% accuracy for campaign classification; Table~\ref{tab:category_acc}). What this means is that not only MetaTroll is able to retain its performance for classifying trolls vs. non-trolls from different campaigns over time, it can also predict which campaign an instance belongs to -- an arguably more difficult task.



\begin{table*}[t]
\small
\caption{Statistics of Multilingual meta-test data. Languages are in ISO code.}
\vspace{-1\baselineskip}
\begin{center}
\begin{tabular}{lccccr@{\;\;}c}
\toprule
\midrule
&Campaign &Event Time &Language & Type & \#users  & Top-3 Hashtags\\ \midrule
&\multirow{2}{*}{Thailand-2020-RTA} &\multirow{2}{*}{Aug 2019 -- Feb 2020} &\multirow{2}{*}{th, en} & Troll & 158 
&\multirow{2}{*}{\shortstack{\#army \#parade \\ \#TheRoyalThaiArmy }} \\
& & & & Non-troll  & 542  \\ \midrule
&\multirow{2}{*}{Mexico-2021-Election} &\multirow{2}{*}{Sep 2020 -- Mar 2021} &\multirow{2}{*}{es, en} & Troll & 190 
&\multirow{2}{*}{\shortstack{\#Elections2021 \\ \#Elections2021MX \#PollsMX}} \\
& & & & Non-troll  & 542  \\ \midrule
&\multirow{2}{*}{Venezuela-2021-Gov} &\multirow{2}{*}{Nov 2020 -- May 2021} &\multirow{2}{*}{es, en} & Troll & 257 
&\multirow{2}{*}{\shortstack{\#Venezuela \\ \#ConTodoPorLaPatria \#Live}} \\
& & & & Non-troll  & 756  \\ \midrule
&\multirow{2}{*}{China-2021-Changyu} &\multirow{2}{*}{Feb 2019 -- Aug 2019} &\multirow{2}{*}{fr,en} & Troll & 280 
&\multirow{2}{*}{\shortstack{\#MBR  \#MorningBox \\ \#StopXinjiangRumors}} \\ 
& & & & Non-troll  & 542  \\ \midrule
&\multirow{2}{*}{Random} &\multirow{2}{*}{Varies} &\multirow{2}{*}{es, fr, th, en} & \multirow{2}{*}{Non-troll} & \multirow{2}{*}{5,000} 
&\multirow{2}{*}{\shortstack{\#news \\ \#COVID \#Daily}} \\ 
& & & & & \\
\midrule
\bottomrule
\end{tabular}
\end{center}
\label{tab:multilingual_data}
\end{table*}


\begin{table}[t]
\small
\caption{\label{tab:multilingual_result} Multilingual results. ``G'' $=$ GRU-2020-NATO, ``I'' $=$ IRA-2020-Russia, ``U'' $=$ Uganda-2021-NRM, ``C'' $=$ China-2021-Xinjiang, ``T'' $=$ Thailand-2020-RTA, ``M'' $=$ Mexico-2021-Election, ``V'' $=$ Venezuela-2021-Gov,``Y'' $=$ China-2021-Changyu}.
\begin{center}
\begin{adjustbox}{max width=1.0\linewidth}
\begin{tabular}{l@{\;\;}c@{\;\;}c@{\;\;}c@{\;\;}c@{\;\;}c@{\;\;}c@{\;\;}c@{\;\;}c}
\toprule
\toprule
Model & G & I & U & C &T  & M & V & Y \\ \toprule
BERT~\cite{conneau2020unsupervised} &51.83 &48.69 &53.21 &55.56 &53.54 &54.98 &54.27 &52.82 \\
KNN~\cite{khandelwalgeneralization} &47.38 &42.50 &44.21 &48.29 &56.36 &51.12 &55.63 &50.19 \\
AdBERT~\cite{pfeiffer2020adapterhub} &58.38 &56.29 &54.79 &66.65 &62.71 &56.20 &58.41 &55.63 \\
\midrule
ProtoNet~\cite{snell2017prototypical} &57.78 &64.47 &60.40 &70.52 &63.68 &59.56 &62.71 &58.96 \\
MAML~\cite{finn2017model} &67.15 &65.58 &55.32 &69.68 &65.32 &57.68 &60.00 &51.36 \\ 
\midrule
MetaTroll &\textbf{69.63} &\textbf{70.52} &\textbf{73.72} &\textbf{73.20} &\textbf{68.14} &\textbf{64.82} &\textbf{67.16} &\textbf{62.53} \\ \bottomrule
\bottomrule 
\end{tabular}
\end{adjustbox}
\end{center}
\vspace{-1\baselineskip}
\end{table}


\subsection{Multilingual performance}
\label{sec:multilingual}
We extend MetaTroll to multilingual. To encode multilingual input text, we replace BERT with XLM-R~\cite{conneau2020unsupervised}.\footnote{Specifically, we use the following library that implements adapters into XLM-R: \url{https://docs.adapterhub.ml/classes/models/xlmroberta.html}} For comparison, we include baselines BERT, KNN, and AdBERT, and meta-learning methods ProtoNet and MAML as their base model can be replaced with XLM-R. We exclude the few-shot classifiers as they are designed for English and cannot be trivially adapted to other languages.

In terms of data, we expand the meta-test campaigns by adding four new campaigns (Thailand-2020-RTA, Mexico-2021-Election, Venezuela-2021-Gov, and China-2021-Changyu) where the predominant language is not in English and present their statistics in Table~\ref{tab:multilingual_data}. For the 10 English campaigns (\tabref{data}), we restore the previously discarded non-English posts and include them for meta-training and meta-testing.

Results are presented in Table~\ref{tab:multilingual_result}.
Generally we see that all models' performance has somewhat degraded when their feature extractor is replaced with a multilingual model, although MetaTroll manages to keep its accuracy around 60\%.
Interestingly, China-2021-Changyu appears to be the most difficult campaign, and we suspect it may be due to its diverse set of languages (38\% French, 12.3\% English, 12.1\% Simplified Chinese).

\begin{table}[t]
\small
\caption{\label{tab:multimodal_result} Multilingual and multimodal results. ``+R'' $=$ ResNet18, ``+O'' $=$ OCR, ``+R+O'' $=$ ``+ResNet18+OCR''.}
\vspace{-1\baselineskip}
\begin{center}
\begin{tabular}{l@{\;\;}c@{\;\;}c@{\;\;}c@{\;\;}c@{\;\;}c@{\;\;}c@{\;\;}c@{\;\;}c}
\toprule
\toprule
Variants  & G & I & U & C &T  & M & V & Y \\\toprule
-image &69.63 &70.52 &73.72 &73.20 &68.14 &64.82 &67.16 &62.53 \\
\midrule
+image\# &70.81 &71.03 &74.02 &73.97 &68.89 &65.11 &67.94 &63.07 \\ 
+R &71.15 &72.81 &75.50 &74.38 &72.25 &67.50 &68.55 &65.74 \\
+O &72.10 &\textbf{74.55} &75.95 &75.59 &71.50 &69.54 &\textbf{72.07} &71.45 \\ 
+R+O &\textbf{73.22} &73.08 &\textbf{76.17} &\textbf{73.04} &\textbf{76.62} &\textbf{70.83} &71.95 &\textbf{72.77} \\ 
\bottomrule
\bottomrule 
\end{tabular}
\end{center}
\vspace{-1\baselineskip}
\end{table}

\subsection{Multimodal performance}
\label{sec:multimodal}
Next we consider incorporating images posted by users, as image is an effective communication device (e.g.\ memes). Note that we only present results for different variants of MetaTroll here, as we have demonstrated that it is the most competitive detection system.

To process images, we use pre-trained ResNet18~\cite{he2016deep} as an off-the-shelf tool to extract image embeddings. We also explore using a multilingual OCR model to extract text information from images (which will be useful for processing memes).\footnote{https://github.com/JaidedAI/EasyOCR} 
As we have multiple images for each user, we aggregate the ResNet18 image embeddings via max-pooling, and concatenate the max-pooled vector with the text representation ($v$ in Equation \ref{eqn:adapterbert}). For texts that are extracted by OCR, we concatenate them into a long string and process them with another AdapterBERT (Equation \ref{eqn:adapterbert}; their parameters are not shared), and similarly concatenate the final CLS representation to the text representation.

Results are in  Table~\ref{tab:multimodal_result}. ``+image\#'' is a baseline where we concatenate a numeric feature that denotes the number of images posted by the user to the text representation. Interestingly, even with the baseline approach we see a small improvement, indicating that trolls use more images (e.g. average number of images used by trolls is 20 vs. 5 for non-trolls in GRU-2020-NATO).
Incorporating either ResNet or OCR encodings boosts performance further (with OCR being marginally more beneficial), and that adding them both produces the best performance.

\section{Conclusion}
We propose MetaTroll, a few-shot troll detection model with
campaign-specific adapters that tackles catastrophic forgetting in a continual learning setting.
Experimental results show that MetaTroll outperforms existing state-of-the-art meta-learning and few-shot text classification models, and it can be extended to handle multilingual and multimodal input.


\section*{Acknowledgements}
This research is supported in part by the Australian Research Council Discovery Project DP200101441.
Lin Tian is supported by the RMIT University Vice-Chancellor PhD Scholarship (VCPS).
\bibliographystyle{ACM-Reference-Format}
\bibliography{troll_www}

\clearpage
\appendix

\section{Implementation Details}
We implement our models in PyTorch using the HuggingFace 
library\footnote{\url{https://github.com/huggingface}} and their 
pretrained BERT\footnote{\url{https://huggingface.co/bert-base-cased}} 
and XML-R\footnote{\url{https://huggingface.co/xlm-roberta-base}}.
The adapter-based models are from AdapterHub\footnote{url{https://github.com/adapter-hub/adapter-transformers}}.

To handle images, we use ResNet18~\footnote{\url{https://pytorch.org/vision/main/models/generated/torchvision.models.resnet18.html}} and EasyOCR~\footnote{\url{https://github.com/JaidedAI/EasyOCR}}.

We set maximum token length $=$ 320 and dropout rate $=$ 0.2 for BERT embedding and dropout rate $=$ 0.1 for meta-learning process. 
Learning rate is tuned in the range between $[1e^{-5}, 5e^{-5}]$ for BERT in the first stage. Learning rate warmup is set up 10\% of steps.
The learning rate $\beta$ for MAML outer loop is $1e^{-5}$.
In the first stage, BERT uses the Adam optimiser~\cite{kingma2014adam}.
Search space for learning rate $\gamma$ and $\delta$ is from $[1e^{-4}, 1e^{-5}, 2e^{-5},3e^{-5},4e^{-5},5e^{-5}]$.
The outer loop learning rate for the third stage $\delta$ is set as $1e^{-5}$.
The inner loop learning rate $\gamma$ for BERT-based models is set as $2e^{-5}$ and for XLM-R-based models is set as $5e^{-5}$.
We experimented with the number of training tasks in the range of 60,000 to 100,000, with 80,000 tasks generally yielding the best results.
Our experiments are running using A100 GPU with 40GB Memory.

\section{Ablation Study}
We conduct ablation study to justify the importance of our three stages training processes. By removing the adaptive classifier, we attach a standard linear layer for the final classification.
For ``s1+s2'', we attach a standard linear layer for the final classification with removing the adaptive classifier. ``s2+s3'' is initialise the BERT text encoder without pre-training on any troll data. ``s1+s3'' drops the meta-trained task-specific adapter. We still allocate one adapter to each task with randomly initialised adaptor. 
We report the average accuracy on four English meta-testing campaigns of different model variants in Table~\ref{tab:ablation_results}. Results show that our model, the combination of all three training stages achieves the best average accuracy of 76.35\%. Without pretraining the base text encoder, the framework performs the worst, resulting in 3.04\% drops on average accuracy.

\begin{table}[tbh]
\caption{\label{tab:ablation_results} Ablation results.}
\begin{center}
\begin{tabular}{l@{\;\;}c@{\;\;}c@{\;\;}c@{\;\;}c@{\;\;}c}
\toprule
\toprule
Variants  & G & I & U & C \\\toprule
 s1 + s2 &71.35 &75.02 &72.47 &80.03 \\ 
 S1 + s3 & 72.43 &74.81 & 72.25 &79.55 \\
 s2 + s3 & 71.25 & 75.19 & 73.50 & 78.33 \\
MetaTroll &72.74 &76.25 &75.05 & 81.37 \\
\bottomrule
\bottomrule 
\end{tabular}
\end{center}
\end{table}

\section{Dataset Details}
\label{sec:appendix_data}
We further include a detailed statistics of our Twitter troll data with detailed topic for each campaign and online post link attached.
\begin{landscape}
\begin{table}[t]
\caption{Statistics of Twitter data for troll and non-troll accounts. Languages are in ISO code.}
\begin{center}
\begin{tabular}{clp{0.52\linewidth}lcrrr}
\toprule
\midrule
&Campaign &Topic & Event Time & Language & \#posts \\ \midrule
&Iran-2018-Palestine & information campaigns potentially originated in Iran, dating back to 2009 [{\href{https://blog.twitter.com/en_us/topics/company/2018/enabling-further-research-of-information-operations-on-twitter}{link}}] &{Feb 2018 -- Aug 2018} & en,fr  & 276,495 \\
&Russia-2016-MAGA & behavior mimics the activity of prior accounts tied to the IRA [{\href{https://blog.twitter.com/en_us/topics/company/2019/further_research_information_operations}{link}}] & {Aug 2015 -- Feb 2016} & en  & 920,761 \\
&Iran-2018-Pakistan & possible malicious activity with an attempted influence campaign identified as potentially located within Iran [{\href{https://blog.twitter.com/en_us/topics/company/2019/further_research_information_operations}{link}}] &  {May 2018 -- Nov 2018} & en & 4,671,959 \\
&Venezuela-2018-Trump & were initial indications that these accounts were associated with the Russian IRA, further analysis suggests that they were operated by a commercial entity originating in Venezuela [{\href{https://blog.twitter.com/en_us/topics/company/2019/information-ops-on-twitter}{link}}] & {Jun 2018 -- Dec 2018} & en & 569,453 \\
&Nigeria-2019-Racism & operating out of Ghana and Nigeria and which we can reliably associate with Russia, attempted to sow discord by engaging in conversations about social issues, like race and civil rights [{\href{https://twitter.com/TwitterSafety/status/1238208545721638912?s=20}{link}}] & {Aug 2019 -- Feb 2020} & en  & 39,964\\
&Thanliand-2020-RTA &a network of accounts partaking in information operations that we can reliably link to the Royal Thai Army (RTA). These accounts were engaging in amplifying pro-RTA and pro-government content, as well as engaging in behavior targeting prominent political opposition figures [\href{https://blog.twitter.com/en_us/topics/company/2020/disclosing-removed-networks-to-our-archive-of-state-linked-information}{link}] & {Aug 2019 -- Feb 2020} & th,en & 21,385 \\
&Iran-2020-BLM & artificially amplified conversations on politically sensitive topics, including Black Lives Matter (BLM), the murder of George Floyd, and other issues of racial and social justice in the United States [\href{https://blog.twitter.com/en_us/topics/company/2020/disclosing-removed-networks-to-our-archive-of-state-linked-information}{link}] & {Jul 2020 -- Jan 2021} & en  & 2,450 \\
&IRA-2020-Russia & accounts amplified narratives that had been previously associated with the IRA and other Russian influence efforts targeting the United States and European Union [{\href{https://blog.twitter.com/en_us/topics/company/2021/disclosing-networks-of-state-linked-information-operations}{link}}] & {Jun 2020 -- Dec 2020} & en & 68,914 \\
&GRU-2020-NATO & accounts amplified narratives that were aligned with the Russian government, while other subset of the network focused on undermining faith in the NATO alliance and its stability  [{\href{https://blog.twitter.com/en_us/topics/company/2021/disclosing-networks-of-state-linked-information-operations}{link}}] & {Jun 2020 -- Dec 2020} & en & 26,684 \\
&Mexico-2021-Election &shared primarily civic content, in support of government initiatives related to public health and political parties.[{\href{https://blog.twitter.com/en_us/topics/company/2021/disclosing-state-linked-information-operations-we-ve-removed}{link}}]  & {Sep 2020 -- Mar 2021} & es,en  & 19,277 \\
&Venezuela-2021-Gov &amplified accounts, hashtags, and topics in support of the government and its official narratives [{\href{https://blog.twitter.com/en_us/topics/company/2021/disclosing-state-linked-information-operations-we-ve-removed}{link}}] & {Nov 2020 -- May 2021} & es,en  & 860,060 \\
&Uganda-2021-NRM &engaged in coordinated inauthentic activity in support of Ugandan presidential incumbent Museveni and his party, National Resistance Movement (NRM) [{\href{https://blog.twitter.com/en_us/topics/company/2021/disclosing-state-linked-information-operations-we-ve-removed}{link}}] & {Jul 2020 -- Jan 2021} & en & 524,081 \\
&China-2021-Xinjiang &amplified Chinese Communist Party narratives related to the treatment of the Uyghur population in Xinjiang [{\href{https://blog.twitter.com/en_us/topics/company/2021/disclosing-state-linked-information-operations-we-ve-removed}{link}}]   & {Jul 2020 -- Jan 2021} & en, zh-cn & 31,269 \\
&China-2021-Changyu &``Changyu Culture'' a private company backed by the Xinjiang regional government [{\href{https://blog.twitter.com/en_us/topics/company/2021/disclosing-state-linked-information-operations-we-ve-removed}{link}}] & {Feb 2019 -- Aug 2019} & fr, en  & 35,924 \\
\midrule
\bottomrule
\end{tabular}
\end{center}
\label{tab:data_full}
\end{table}
\end{landscape}

\end{document}